\documentclass[sigconf]{acmart}

\def\BibTeX{{\rm B\kern-.05em{\sc i\kern-.025em b}\kern-.08emT\kern-.1667em\lower.7ex\hbox{E}\kern-.125emX}}

\usepackage{nicefrac}
\usepackage{siunitx}
\usepackage{array,framed}
\usepackage{booktabs}
\usepackage{
  color,
  float,
  epsfig,
  wrapfig,
  graphics,
  graphicx,
  subcaption
}
\usepackage{textcomp}
\usepackage{setspace}
\usepackage{latexsym,fancyhdr,url}
\usepackage{enumerate}
\usepackage{algorithm2e}
\usepackage{algpseudocode}
\usepackage{xparse} 
\usepackage{xspace}
\usepackage{multirow}
\usepackage{csvsimple}
\usepackage{balance}

\setcopyright{acmcopyright}\acmConference[SenSys '20]{The 18th ACM Conference on Embedded Networked Sensor Systems}{November 16--19, 2020}{Virtual Event, Japan}
\copyrightyear{2020}
\acmYear{2020}
\acmPrice{15.00}
\acmDOI{10.1145/3384419.3430421}
\acmISBN{978-1-4503-7590-0/20/11}

\usepackage{dblfloatfix}

\usepackage{
  tikz,
  pgfplots,
  pgfplotstable
}
\usepackage{hyperref}

\usetikzlibrary{
  shapes.geometric,
  arrows,
  external,
  pgfplots.groupplots,
  matrix
}

\pgfplotsset{compat=1.9}

\usepackage{mathtools,}

\DeclareMathAlphabet{\mathcal}{OMS}{cmsy}{m}{n}

\usepackage{xparse}
\newcommand{\bnm}{\begin{newmath}}
\newcommand{\enm}{\end{newmath}}

\newcommand{\bea}{\begin{eqnarray*}}%
\newcommand{\eea}{\end{eqnarray*}}%

\newcommand{\bne}{\begin{newequation}}
\newcommand{\ene}{\end{newequation}}

\newcommand{\bal}{\begin{newalign}}
\newcommand{\eal}{\end{newalign}}

\newenvironment{newalign}{\begin{align}%
\setlength{\abovedisplayskip}{4pt}%
\setlength{\belowdisplayskip}{4pt}%
\setlength{\abovedisplayshortskip}{6pt}%
\setlength{\belowdisplayshortskip}{6pt} }{\end{align}}

\newenvironment{newmath}{\begin{displaymath}%
\setlength{\abovedisplayskip}{4pt}%
\setlength{\belowdisplayskip}{4pt}%
\setlength{\abovedisplayshortskip}{6pt}%
\setlength{\belowdisplayshortskip}{6pt} }{\end{displaymath}}

\newenvironment{newequation}{\begin{equation}%
\setlength{\abovedisplayskip}{4pt}%
\setlength{\belowdisplayskip}{4pt}%
\setlength{\abovedisplayshortskip}{6pt}%
\setlength{\belowdisplayshortskip}{6pt} }{\end{equation}}

\newcounter{ctr}

\newcounter{mytable}
\def\mytable{\begin{centering}\refstepcounter{mytable}}
\def\endmytable{\end{centering}}

\newcounter{myfig}
\def\myfig{\begin{centering}\refstepcounter{myfig}}
\def\endmyfig{\end{centering}}

\newlength{\saveparindent}
\newlength{\saveparskip}
\newcommand{\E}{{\rm I\kern-.3em E}}

\newcommand{\figref}[1]{\mbox{Figure~\ref{#1}}}

\renewcommand{\eqref}[1]{\mbox{Equation~(\ref{#1})}}

\def \part {part}

\renewcommand{\paragraph}[1]{\vspace*{6pt}\noindent\textbf{#1}\;}

\def \blackslug{\hbox{\hskip 1pt \vrule width 4pt height 8pt
    depth 1.5pt \hskip 1pt}}
\def \qed{\quad\blackslug\lower 8.5pt\null\par}

\newcounter{mynote}[section]

\newcommand\ignore[1]{}

\newcounter{rcnote}[section]

\newcounter{mrnote}[section]

\newcounter{fknote}[section]

\newcounter{anote}[section]

\DeclareMathSymbol{\mlq}{\mathord}{operators}{``}
\DeclareMathSymbol{\mrq}{\mathord}{operators}{`'}

\newcommand{\rhf}[2]{R_{f, \gamma}}

\DeclareDocumentCommand{\edist}{o o}{
  \ensuremath{
    \IfNoValueTF{#1}{{d}}{{\sf d}(#1,#2)}
  }
}

\newcommand{\olrk}[1]{\ifx\nursymbol#1\else\!\!\mskip4.5mu plus 0.5mu\left(\mskip0.5mu plus0.5mu #1\mskip1.5mu plus0.5mu \right)\fi}

\NewDocumentCommand{\indseq}{ O{1} O{r} }{{#1}\ldots {#2}}

\begin{document}
\fancyhead{}
\def\thetitle{Demo Abstract: Indoor Positioning System in Visually-Degraded Environments with Millimetre-Wave Radar and Inertial Sensors}
\title{\thetitle}

\author{
Zhuangzhuang Dai$^{\ast,1}$, Muhamad Risqi U. Saputra$^{\ast,1}$, Chris Xiaoxuan Lu$^{\ast,2}$} 

\author{Niki Trigoni$^{1}$, Andrew Markham$^{1}$
}

\affiliation{%
  \institution{$^{1}$ University at Oxford, Oxford, England, United Kingdom}
  \institution{$^{2}$ University of Edinburgh, Edinburgh, Scotland, United Kingdom}
  \institution{$^\ast$ Equal contribution.}
}

\begin{abstract}
Positional estimation is of great importance in the public safety sector. Emergency responders such as fire fighters, medical rescue teams, and the police will all benefit from a resilient positioning system to deliver safe and effective emergency services. Unfortunately, satellite navigation (e.g., GPS) offers limited coverage in indoor environments. It is also not possible to rely on infrastructure based solutions. To this end, wearable sensor-aided navigation techniques, such as those based on camera and Inertial Measurement Units (IMU), have recently emerged recently as an accurate, infrastructure-free solution. Together with an increase in the computational capabilities of mobile devices, motion estimation can be performed in real-time. In this demonstration, we present a real-time indoor positioning system which fuses millimetre-wave (mmWave) radar and IMU data via deep sensor fusion. We employ mmWave radar rather than an RGB camera as it provides better robustness to visual degradation (e.g., smoke, darkness, etc.) while at the same time requiring lower computational resources to enable runtime computation. We implemented the sensor system on a handheld device and a mobile computer running at 10 FPS to track a user inside an apartment. Good accuracy and resilience were exhibited even in poorly illuminated scenes.

\end{abstract}
\maketitle

\section*{CCS CONCEPTS}
\label{sec:ccs}
• Computer systems organization → Real-time systems; • Human-centered computing → Ubiquitous and mobile computing; • Computing methodologies → Artificial intelligence; Machine learning.
\section*{keywords}
Millimeter-wave sensor, IMU, Deep Learning, Indoor Positioning.

\section*{\small{ACM Reference Format}}
Zhuangzhuang Dai, Muhamad Risqi U. Saputra, Chris Xiaoxuan Lu, Niki Trigoni, and Andrew Markham. 2020. Demo Abstract: Real-Time Positioning System in Visually-Degraded Environments with Millimetre-Wave Radar and Inertial Sensors. In Proceedings of The 18th ACM International Conference on Embedded Networked Sensor Systems (SenSys), Nov 16-19, 2020, Virtual Event, Japan.

\section{Introduction}
\label{sec:intro}

There is a long-standing and unsolved need to be able to locate emergency responders in indoor environments. Since satellite and radio positioning techniques perform poorly indoors, sensor-aided navigation such as that based on Visual-Inertial Odometry (VIO) has been increasingly adopted. State-of-the-art VIO algorithms have seen great success in tracking humans and robots \cite{ref1, ref2}. However, the performance of camera sensors \cite{ref3} is impacted by visual degradation such as from glare, darkness, or smoke. Although an IMU is characterized as being egocentric and environment-agnostic sensor, it cannot perform reliable tracking on its own due noise and intrinsic bias \cite{ref4}.
\begin{figure}[b]
	\centering
	\includegraphics[width=0.44\textwidth]{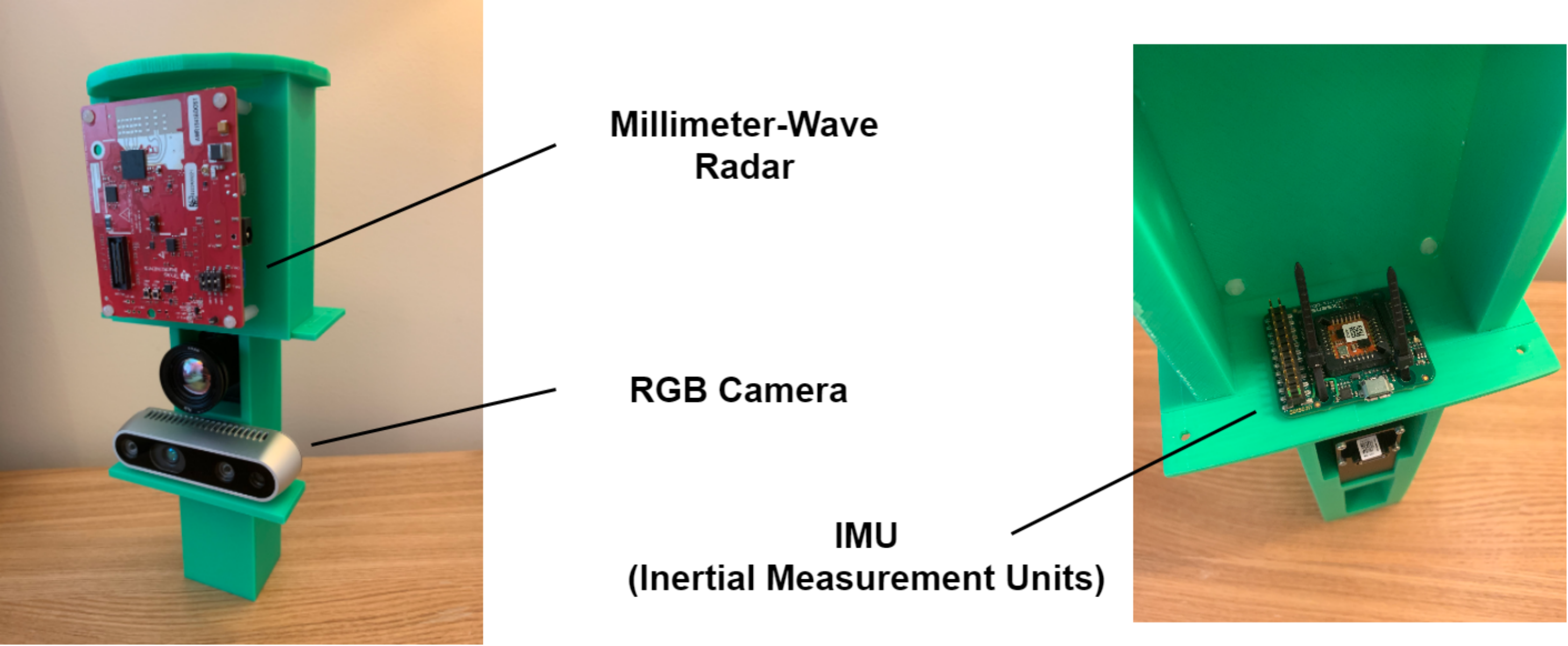}
	\caption[fig1]{Handheld device that holds a TI AWR1843 mmWave radar and an Xsens MTi 1-series IMU. It also contains a Flir Boson thermal camera, and an Intel D435i Depth camera which are used for ground-truthing.}
	\label{sensors}
\end{figure}

In this work, we propose to use mmWave radar, a low-cost, low-power, and lightweight sensor, in conjunction with the IMU to track first-responders in visually-degraded environments. The hypothesis is that the egocentric motions estimated by IMU can be significantly improved with sparse scene sensing from mmWave radar. The mmWave radar emits FMCW chirps  measures the time-of-flight of reflected signals. The 3-by-4 antennas array exploits the phase difference to determine the angle of arrival. We convert the radar measurements into depth images for two-view odometry analysis. Note that these depth images are far sparser than RGB due to the limited sensor resolution. Given consecutive mmWave depth images and IMU sequences, a DNN \cite{lu2020milliego} extracts optical flow-like features from the sensors and uses them to estimate the camera ego motion. We 3D printed a handheld device to hold the mmWave radar and the IMU as shown in \figref{sensors}. A thermal camera and an RGB camera are also installed for trajectory visualization but are not used in the calculations. Thanks to the low data-rate of the mmWave sensor (in the order of 1000 points/second), poses of the handheld device can be calculated at a high output rate.



\section{MMWave-Inertial Odometry}
\label{sec:relwork}

\begin{figure}
	\centering
	\includegraphics[width=0.45\textwidth]{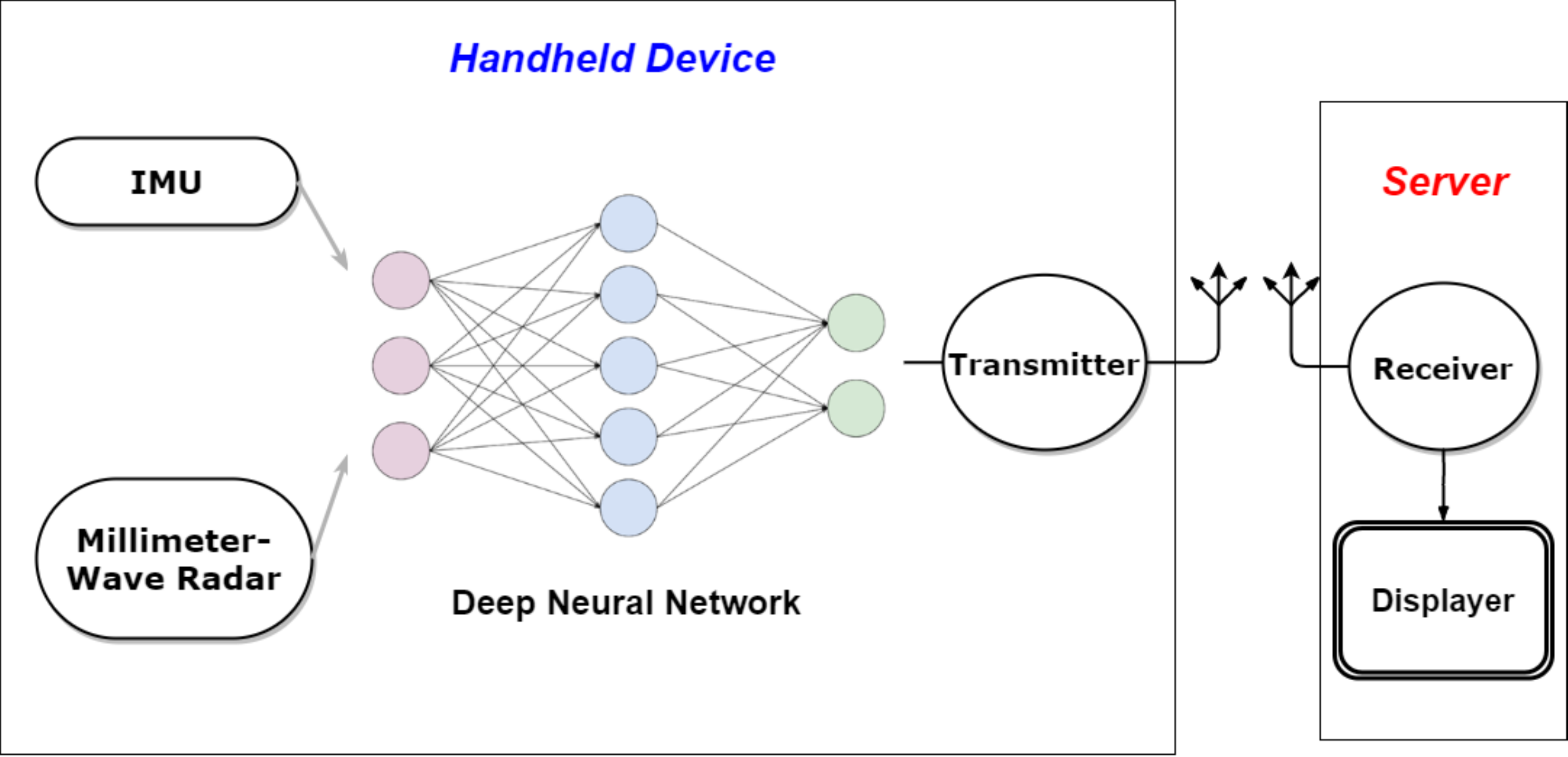}
	\caption[fig3]{System architecture: IMU and mmWave radar are fused before invoking the DNN inference engine; Obtained poses are transmitted to the server over the air.}
	\label{sys}
\end{figure}

The point clouds obtained from mmWave radar are highly noisy due to diffuse scattering. Since the angular resolution of the antenna array is limited, the point cloud images can also be very sparse \cite{ref5}. In order to combat the irregularities from mmWave radar, we overlay three consecutive point clouds and project them into one panoramic-view image before normalizing the values. A CNN feature extractor is used to capture mmWave features which are further fused with inertial features obtained from IONET \cite{ref6}. A mixed attention mechanism \cite{lu2020milliego} is then employed to selectively fuse the most important features for ego motion estimation. On receiving the fused mmWave-inertial features, an RNN is used to model the long-term motion dynamics. Finally, three fully connected (FC) layers are introduced to regress relative motion transformation.

There is a practical challenge to run the model at runtime. The DNN model has 33.9 million parameters which takes 187 MB of storage. It takes a 4th generation Raspberry Pi 0.5s to generate a single inference. Instead, we use NVIDIA Jetson AGX Xavier instead to prototype a capable mobile computer which supports GPU and Deep Learning Accelerating engines in an embedded module under 30 W. The Xavier delivers an average inference latency of 0.034s, i.e., over 20 FPS processing rate. We use Robot Operating System (ROS) for sensor synchronization and data collection. ROS is a \textit{de-facto} standard in robotics which allows easy system integration. \figref{sys} exhibits a decomposition of the runtime sensor fusion and data processing system.

\section{Demonstration}
\label{sec:methodology}

\begin{figure}[!ht]
	\centering
	\begin{subfigure}[b]{0.23\textwidth}\centering
		\includegraphics[width=\columnwidth]{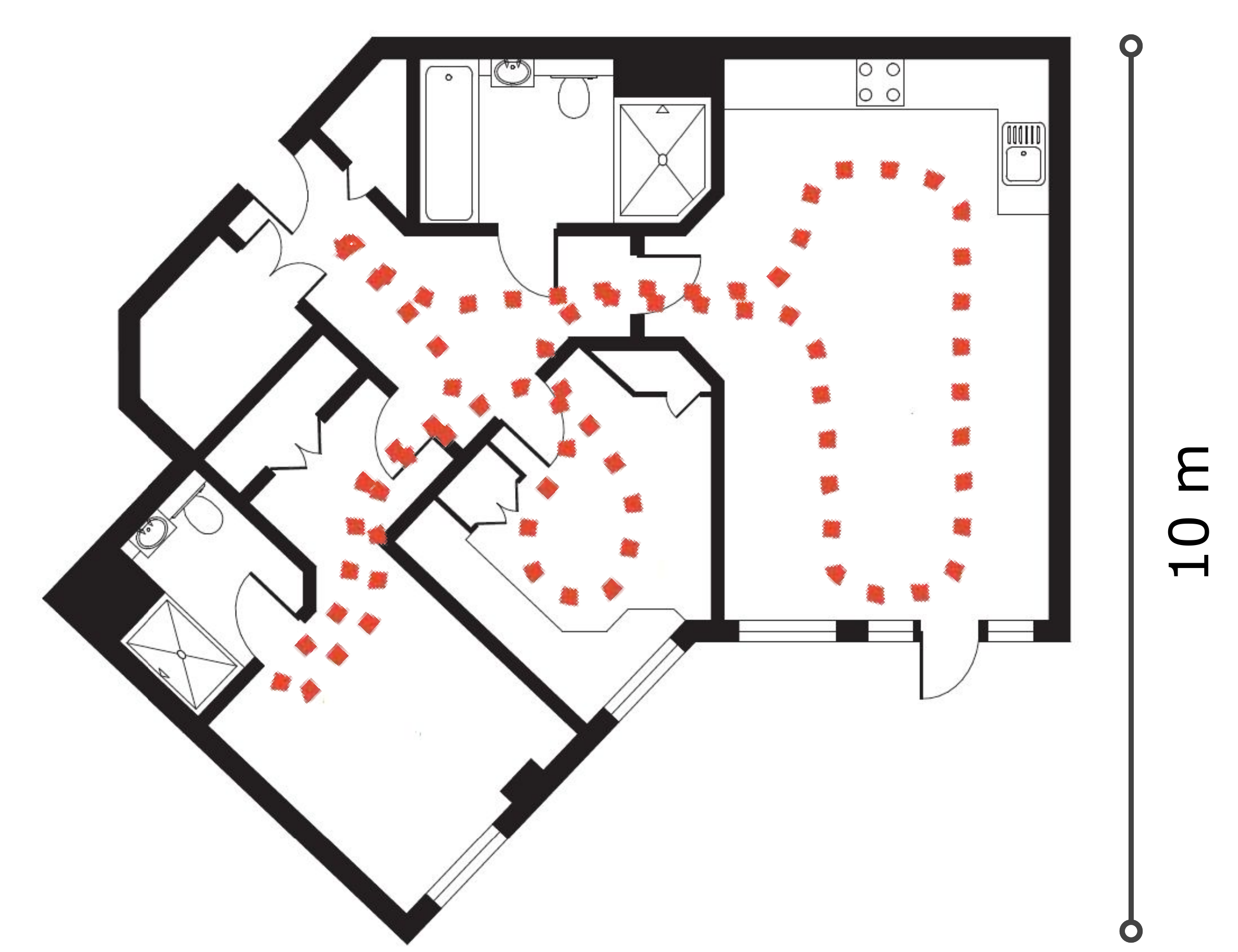} 
		\caption{True trajectory (red)}
	\end{subfigure}%
	\hfill
	\begin{subfigure}[b]{0.23\textwidth}\centering
		\includegraphics[width=\columnwidth]{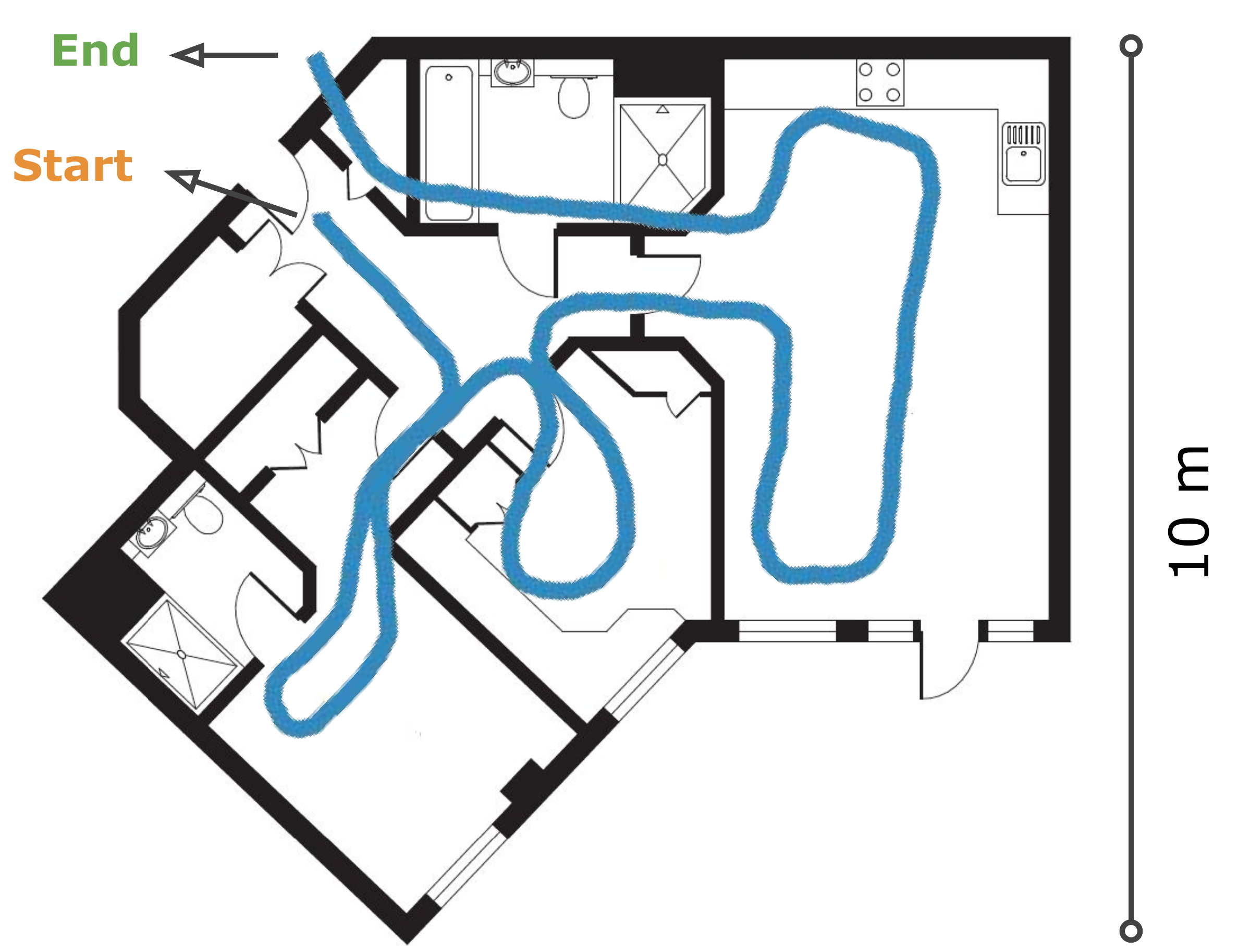} 
		\caption{Estimation (blue)}
	\end{subfigure}%
	\vspace{5mm}
\caption[fig3]{Trajectory of a search procedure carried out in a two-bed apartment: the ground-truth path is marked in a dashed red line in (a); the estimated path is marked in a solid blue line in (b).}
\label{fig:generalization_dai}
\end{figure}

In this demonstration, we present the performance of the positioning systems for indoor tracking. The demonstration shows a person performing search procedures (used by firefighters in UK) throughout an apartment including a dark bedroom. Trajectory of a completed search is illustrated in \figref{fig:generalization_dai}. The user is taking a battery-powered handheld device comprised of an NVIDIA Jetson AGX Xavier developer kit, a mmWave radar, and an IMU to deliver runtime computation and recording. Presented screens include the calculated poses from sensors, the ground-truth path generated by a Velodyne LiDAR, and a panoramic view of the point clouds from the mmWave radar. Excellent accuracy and reliability are seen even in poorly illuminated scenes.


\section*{Acknowledgement}
\label{sec:eval}

This research has been financially supported by the National Institute of Standards and Technology (NIST) via the grant Pervasive, Accurate, and Reliable Location-based Services for Emergency Responders (Federal Grant: 70NANB17H185).

\bibliographystyle{ACM-Reference-Format}
\bibliography{bib}


\begin{thebibliography}{00}


\ifx \showCODEN    \undefined \def \showCODEN     #1{\unskip}     \fi
\ifx \showDOI      \undefined \def \showDOI       #1{#1}\fi
\ifx \showISBNx    \undefined \def \showISBNx     #1{\unskip}     \fi
\ifx \showISBNxiii \undefined \def \showISBNxiii  #1{\unskip}     \fi
\ifx \showISSN     \undefined \def \showISSN      #1{\unskip}     \fi
\ifx \showLCCN     \undefined \def \showLCCN      #1{\unskip}     \fi
\ifx \shownote     \undefined \def \shownote      #1{#1}          \fi
\ifx \showarticletitle \undefined \def \showarticletitle #1{#1}   \fi
\ifx \showURL      \undefined \def \showURL       {\relax}        \fi
\providecommand\bibfield[2]{#2}
\providecommand\bibinfo[2]{#2}
\providecommand\natexlab[1]{#1}
\providecommand\showeprint[2][]{arXiv:#2}

\bibitem[\protect\citeauthoryear{Chen, Lu, Markham, and Trigoni}{Chen
  et~al\mbox{.}}{2018}]%
        {ref4}
\bibfield{author}{\bibinfo{person}{C. Chen}, \bibinfo{person}{X. Lu},
  \bibinfo{person}{A. Markham}, {and} \bibinfo{person}{N. Trigoni}.}
  \bibinfo{year}{2018}\natexlab{}.
\newblock \showarticletitle{IONet: Learning to Curethe Curse of Drift in
  Inertial Odometry}. In \bibinfo{booktitle}{{\em Association for the
  Advancement of Artificial Intelligence}}.
\newblock


\bibitem[\protect\citeauthoryear{Chen, Rosa, Miao, Lu, Wu, Markham, and
  Trigoni}{Chen et~al\mbox{.}}{2019}]%
        {ref2}
\bibfield{author}{\bibinfo{person}{C. Chen}, \bibinfo{person}{S. Rosa},
  \bibinfo{person}{Y. Miao}, \bibinfo{person}{C.~X. Lu}, \bibinfo{person}{W.
  Wu}, \bibinfo{person}{A. Markham}, {and} \bibinfo{person}{N. Trigoni}.}
  \bibinfo{year}{2019}\natexlab{}.
\newblock \showarticletitle{Selective sensor fusion for neural visual-inertial
  odometry}. In \bibinfo{booktitle}{{\em Proceedings of IEEE Conference on
  Computer Vision and Pattern Recognition}}.
\newblock


\bibitem[\protect\citeauthoryear{Chen, Zhao, Lu, Wang, Markham, and
  Trigoni}{Chen et~al\mbox{.}}{2020}]%
        {ref6}
\bibfield{author}{\bibinfo{person}{C. Chen}, \bibinfo{person}{P. Zhao},
  \bibinfo{person}{C.~X. Lu}, \bibinfo{person}{W. Wang}, \bibinfo{person}{A.
  Markham}, {and} \bibinfo{person}{N. Trigoni}.}
  \bibinfo{year}{2020}\natexlab{}.
\newblock \showarticletitle{Deep Learning based Pedestrian Inertial Navigation:
  Methods, Dataset and On-Device Inference}. In \bibinfo{booktitle}{{\em IEEE
  Internet of Things Journal}}.
\newblock


\bibitem[\protect\citeauthoryear{Leutenegger, Lynen, Bosse, Siegwart, and
  Furgale}{Leutenegger et~al\mbox{.}}{2015}]%
        {ref1}
\bibfield{author}{\bibinfo{person}{S. Leutenegger}, \bibinfo{person}{S. Lynen},
  \bibinfo{person}{M. Bosse}, \bibinfo{person}{R. Siegwart}, {and}
  \bibinfo{person}{P. Furgale}.} \bibinfo{year}{2015}\natexlab{}.
\newblock \showarticletitle{Keyframe-based visual–inertial odometry using
  nonlinear optimization}. In \bibinfo{booktitle}{{\em The International
  Journal of Robotics Research 34, 3 (2015), 314–334}}.
\newblock


\bibitem[\protect\citeauthoryear{Lu, Saputra, Zhao, Almalioglu, de~Gusmao,
  Chen, Sun, Trigoni, and Markham}{Lu et~al\mbox{.}}{2020}]%
        {lu2020milliego}
\bibfield{author}{\bibinfo{person}{C.~X. Lu}, \bibinfo{person}{M.~R.~U
  Saputra}, \bibinfo{person}{P. Zhao}, \bibinfo{person}{Y. Almalioglu},
  \bibinfo{person}{P.~P.~B. de Gusmao}, \bibinfo{person}{C. Chen},
  \bibinfo{person}{K. Sun}, \bibinfo{person}{N. Trigoni}, {and}
  \bibinfo{person}{A. Markham}.} \bibinfo{year}{2020}\natexlab{}.
\newblock \showarticletitle{milliEgo: Single-chip mmWave Radar Aided Egomotion
  Estimation via Deep Sensor Fusion}. In \bibinfo{booktitle}{{\em ACM
  Conference on Embedded Networked Sensor Systems (SenSys)}}.
\newblock


\bibitem[\protect\citeauthoryear{Saputra, de~Gusmao, Lu, Almalioglu, Rosa,
  Chen, Wahlstr{\"o}m, Wang, Markham, and Trigoni}{Saputra
  et~al\mbox{.}}{2020}]%
        {ref3}
\bibfield{author}{\bibinfo{person}{M.~R.~U. Saputra}, \bibinfo{person}{P.~P.~B.
  de Gusmao}, \bibinfo{person}{C.~X. Lu}, \bibinfo{person}{Y. Almalioglu},
  \bibinfo{person}{S. Rosa}, \bibinfo{person}{C. Chen}, \bibinfo{person}{J.
  Wahlstr{\"o}m}, \bibinfo{person}{W. Wang}, \bibinfo{person}{A. Markham},
  {and} \bibinfo{person}{N. Trigoni}.} \bibinfo{year}{2020}\natexlab{}.
\newblock \showarticletitle{Deeptio: A deep thermal-inertial odometry with
  visual hallucination}.
\newblock \bibinfo{journal}{{\em IEEE Robotics and Automation Letters\/}}
  \bibinfo{volume}{5}, \bibinfo{number}{2} (\bibinfo{year}{2020}),
  \bibinfo{pages}{1672--1679}.
\newblock


\bibitem[\protect\citeauthoryear{Zhao, Lu, Wang, Chen, Wang, Trigoni, and
  Markham}{Zhao et~al\mbox{.}}{2019}]%
        {ref5}
\bibfield{author}{\bibinfo{person}{P. Zhao}, \bibinfo{person}{C.~X. Lu},
  \bibinfo{person}{J. Wang}, \bibinfo{person}{C. Chen}, \bibinfo{person}{W.
  Wang}, \bibinfo{person}{N. Trigoni}, {and} \bibinfo{person}{A. Markham}.}
  \bibinfo{year}{2019}\natexlab{}.
\newblock \showarticletitle{mID: Tracking and Identifying People with
  Millimeter WaveRadar}. In \bibinfo{booktitle}{{\em 15th International
  Conference on Distributed Computing in SensorSystems (DCOSS)}}.
\newblock


\end{thebibliography}


\end{document}